\documentclass[11pt,a4paper]{article}
\usepackage[hyperref]{acl2020}
\usepackage{times}
\usepackage{latexsym}
\usepackage{tabularx}
\usepackage{amsmath}
\usepackage{booktabs}

\usepackage{microtype}

\aclfinalcopy 

\newcommand{\pgen}{p_{\text{gen}}}
\newcommand{\pvocab}{P_{\text{vocab}}}
\newcommand{\pcopy}{P_{\text{copy}}}
\usepackage{graphicx}
\graphicspath{ {./figures/} }
\usepackage{multirow}
\usepackage{amssymb}

\title{The NYU-CUBoulder Systems for\\SIGMORPHON 2020 Task 0 and Task 2}

\author{Assaf Singer \\
  New York University \\
  USA \\
  \texttt{as12152@nyu.edu} \\\And
  Katharina Kann \\
  University of Colorado Boulder \\
  USA \\
  \texttt{katharina.kann@colorado.edu} \\}

\date{}

\begin{document}
\maketitle
\begin{abstract}

We describe the NYU-CUBoulder systems for the SIGMORPHON 2020 Task 0 on typologically diverse morphological inflection and Task 2 on unsupervised morphological paradigm completion.
The former consists of generating morphological inflections from a lemma and a set of morphosyntactic features describing the target form. The latter requires generating entire paradigms for a set of given lemmas from raw text alone.
We model morphological inflection as a sequence-to-sequence problem, where the input is the sequence of the lemma's characters with morphological tags, and the output is the sequence of the inflected form's characters. 
First, we apply a transformer model to the task.
Second, as inflected forms share most characters with the lemma, we further propose a pointer-generator transformer model to allow easy copying of input characters. Our best performing system for Task 0 is placed 6th out of 23 systems.
We further use our inflection systems as subcomponents of approaches for Task 2. Our best performing system for Task 2 is the 2nd best out of 7 submissions.

\end{abstract}

\section{Introduction}
In morphologically rich languages, a word’s surface form reflects syntactic and semantic properties that are expressed by the word. For example, most English nouns have both singular and plural forms (e.g., \textit{robot}/\textit{robots}, \textit{process}/\textit{processes}), which are known as the inflected forms of the noun. Some languages display little inflection. In contrast, others have many inflections per base form or lemma: a Polish verb has nearly 100 inflected forms \citep{Janecki1998} and an Archi verb has around 1.5 million \citep{kibrik1998archi}. 

Morphological inflection is the task of, given an input word -- a lemma -- together with morphosyntactic features defining the target form, generating the indicated inflected form, cf. Figure \ref{fig:morphexample}.  Morphological inflection is a useful tool for many natural language processing tasks \citep{DBLP:journals/tacl/SeekerC15, DBLP:conf/acl/CotterellSE16}, especially in morphologically rich languages where handling inflected forms can reduce data sparsity \citep{DBLP:conf/acl/MinkovTS07}.

\begin{figure}[t!]
\centering
\begin{tabular}{lll}
\textbf{Lemma} & \textbf{Features} & \textbf{Inflected form} \\
hug & V;PST & hugged \\
seel & V;3;SG;PRS & seels
\end{tabular}
\caption{Morphological inflection examples in English. A lemma and features are mapped to an inflected form.}
\label{fig:morphexample}
\end{figure}

The SIGMORPHON 2020 Shared Task consists of three separate tasks. We participate in Task 0 on typologically diverse morphological inflection \cite{vylomova2020sigmorphon} and Task 2 on unsupervised morphological paradigm completion \cite{kann-etal-2020-sigmorphon}. 
\textbf{Task 0} consists of generating morphological inflections from a lemma and a set of morphosyntactic features describing the target form.
For this task, we implement a pointer-generator transformer model, based on the vanilla transformer model \citep{DBLP:conf/nips/VaswaniSPUJGKP17} and the pointer-generator model \cite{DBLP:conf/acl/SeeLM17}. After adding a copy mechanism to the transformer, it produces a final probability distribution as a combination of generating elements from its output vocabulary and copying elements -- characters in our case -- from the input. 
As most inflected forms derive their characters from the source lemma, the use of a mechanism for copying characters directly from the lemma has proven to be effective for morphological inflection generation, especially in the low resource setting \citep{DBLP:conf/acl/AharoniG17,DBLP:journals/corr/MakarovRC17}.

For our submissions, we further increase the size of all training sets by performing multi-task training on morphological inflection and morphological reinflection, i.e., the task of generating inflected forms from forms \textit{different from the lemma}. 
For languages with small training sets, we also perform hallucination pretraining \citep{DBLP:conf/emnlp/AnastasopoulosN19}, where we generate pseudo training instances for the task, based on suffixation and prefixation rules collected from the original dataset.

For \textbf{Task 2}, participants are given raw text and a source file with lemmas. The objective is to generate the complete paradigms for all lemmas. Our systems for this task consist of a combination of the official baseline system \citep{jin2020unsupervised} and our systems for Task 0. The baseline system finds inflected forms in the text, decides on the number of inflected forms per lemma, and produces pseudo training files for morphological inflection. Our inflection model then learns from these and, subsequently, generates all missing forms. 

\section{Related Work}

\paragraph{SIGMORPHON and CoNLL--SIGMORPHON shared tasks.}
In recent years, the SIGMORPHON and CoNLL–SIGMORPHON shared tasks have promoted research on computational morphology, with a strong focus on morphological inflection.
Research related to those shared tasks includes \citet{DBLP:conf/acl/KannS16}, who used an LSTM \citep{DBLP:journals/neco/HochreiterS97} sequence-to-sequence model with soft attention \citep{DBLP:journals/corr/BahdanauCB14} and achieved the best result in the SIGMORPHON 2016 shared task \citep{DBLP:conf/sigmorphon/KannS16, DBLP:conf/sigmorphon/CotterellKSYEH16}. Due to the often monotonic alignment between input and output, \citet{DBLP:conf/acl/AharoniG17} proposed a model with hard monotonic attention. Based on this, \citet{DBLP:journals/corr/MakarovRC17} implemented a neural state-transition system which also used hard monotonic attention and achieved the best results for Task 1 of the SIGMORPHON 2017 shared task. 
In 2018, the best results were achieved by a revised version of the neural transducer, trained with imitation learning \citep{DBLP:conf/conll/MakarovC18}. That model learned an alignment instead of maximizing the likelihood of gold action sequences given by a separate aligner.

\paragraph{Transformers.}
Transformers have produced state-of-the-art results on various tasks such as machine translation \citep{DBLP:conf/nips/VaswaniSPUJGKP17} language modeling \citep{DBLP:conf/aaai/Al-RfouCCGJ19}, question answering \citep{DBLP:conf/naacl/DevlinCLT19} and language understanding \citep{DBLP:conf/naacl/DevlinCLT19}.
There has been very little work on transformers for morphological inflection, with, to the best of our knowledge, \citet{DBLP:journals/corr/abs-2005-01630} being the only published paper. However, the widespread success of transformers in NLP leads us to believe that a transformer model could perform well on morphological inflection.

\paragraph{Pointer-generators.}
In addition to the transformer, the architecture of our model is also inspired by \citet{DBLP:conf/acl/SeeLM17}, who used a pointer-generator network for abstractive summarization. Their model could choose between generating a new element and copying an element from the input directly to the output. This copying of words from the source text via pointing \citep{DBLP:conf/nips/VinyalsKKPSH15}, improved the handling of out-of-vocabulary words. Copy mechanisms have also been used for other tasks, including morphological inflection \citep{DBLP:conf/conll/SharmaKS18}. \textit{Transformers} with copy mechanisms have been used for word-level tasks \citep{DBLP:conf/naacl/ZhaoWSJL19}, but, as far as we know, never before on the character level.

\section{SIGMORPHON 2020 Shared Task}
The SIGMORPHON 2020 Shared Task is composed of three tasks: Task 0 on typologically diverse morphological inflection \citep{vylomova2020sigmorphon}, Task 1 on multilingual grapheme-to-phoneme conversion \citep{Task1-2020-sigmorphon}, and Task 2 on unsupervised morphological paradigm completion \citep{kann-etal-2020-sigmorphon}. We submit systems to Tasks 0 and 2.

\subsection{Task 0: Typologically Diverse Morphological Inflection}
SIGMORPHON 2020 Task 0 focuses on morphological inflection in a set of typologically diverse languages. Different languages inflect differently, so it is not trivially clear that systems that work on some languages also perform well on others.
For Task 0, systems need to generalize well to a large group of languages, including languages unseen during model development.

The task features 90 languages in total. 45 of them are development languages, coming from five families: Austronesian, Niger–Congo, Uralic, Oto-Manguean, and Indo-European. The remaining 45 are surprise languages, and many of those are from language families different from the development languages. Some languages have very small training sets, which makes them hard to model. For those cases, the organizers recommend a family-based multilingual approach to exploit similarities between related languages. While this might be effective, we believe that using multitask training in combination with hallucination pretraining can give the model enough information to learn the task well, while staying true to the specific structure of each individual language.

\subsection{Task 2: Unsupervised Morphological Paradigm Completion}
Task 2 is a novel task, designed to encourage work on unsupervised methods for computational morphology. 
As morphological annotations are limited for many of the world's languages, the study of morphological generation in the low-resource setting is of great interest \citep{DBLP:conf/conll/CotterellKSWVMK18}. However, a different way to tackle the problem is by creating systems that are able to use data without annotations.

For Task 2, a tokenized Bible in each language is given to the participants, along with a list of lemmas. Participants should then produce complete paradigms for each lemma.
As slots in the paradigm are not labeled with gold data paradigm slot descriptions, an evaluation metric called best-match accuracy was designed for this task.
First, this metric matches predicted paradigm slots with gold slots in the way which leads to the highest overall accuracy. It then evaluates the correctness of individual inflected forms.

\section{Methods}
In this section, we introduce our models for Tasks 0 and 2 and describe all approaches we use, such as multitask training, hallucination pretraining and ensembling.
The code for our models is available online.\footnote{\href{https://github.com/AssafSinger94/sigmorphon-2020-inflection}{github.com/AssafSinger94/sigmorphon-2020-inflection}}

\subsection{Transformer}
Our model is built on top of the transformer architecture \citep{DBLP:conf/nips/VaswaniSPUJGKP17}. It consists of an encoder and a decoder, each composed of a stack of layers. Each encoder layer consists, in turn, of a self-attention layer, followed by a fully connected layer. Decoder layers contain an additional inter-attention layer between the two.

With inputs $\left( x_1, \cdots, x_T \right)$ being a lemma's characters followed by tags representing the morphosyntactic features of the target form, the encoder processes the input sequence and outputs hidden states $\left( h_1, \cdots, h_T \right)$.
At generation step $t$, the decoder reads the previously generated sequence $\left( y_1, \cdots, y_{t-1} \right)$ to produce states $\left( s_1, \cdots, s_{t-1} \right)$. 
The last decoder state $s_{t-1}$ is then passed through a linear layer followed by a softmax, to generate a probability distribution over the output vocabulary:
\begin{equation}
    \pvocab = \textrm{softmax}(V s_{t-1} + b)
\end{equation}
During training, the entire target sequence $\left( y_1, \cdots, y_{T_y} \right)$ is input to the decoder at once, along with a sequential mask to prevent positions from attending to subsequent positions.

\subsection{Pointer-Generator Transformer}
The pointer-generator transformer allows for both generating characters from a fixed vocabulary, as well as copying from the source sequence via pointing \citep{DBLP:conf/nips/VinyalsKKPSH15}. This is managed by $\pgen$ -- the probability of generating as opposed to copying -- which acts as a soft switch between the two actions. $\pgen$ is computed by passing a concatenation of the decoder state $s_{t}$, the previously generated output $y_{t-1}$, and a context vector $c_t$ through a linear layer, followed by the sigmoid function.
\begin{equation}
    \pgen = \sigma(w [s_t;c_t;y_{t-1}] + b)
\end{equation}
The context vector is computed as the weighted sum of the encoder hidden states
\begin{equation}
    c_t = \sum\nolimits_{i=1}^{T} a^t_i h_i
\end{equation}
with attention weights $\left( a^t_1, \cdots, a^t_T \right)$.
For each inflection example, let the extended vocabulary denote the union of the output vocabulary, and all characters appearing in the source lemma.
We then use $\pgen$, $\pvocab$ produced by the transformer, and the attention weights of the last decoder layer $\left( a^t_1, \cdots, a^t_T \right)$ to compute a distribution over the extended vocabulary:
\begin{equation}
    P(c) = \pgen \pvocab(c) + (1-\pgen) \pcopy(c),
\end{equation}
with
\begin{equation}
    \pcopy(c) = \sum\nolimits_{i: x_i = c} a^t_i
\end{equation}
The copy distribution $\pcopy(c)$ for each character $c$ is the sum of attention weights over all source positions where $x_i = c$.
Note that if $c$ is an out-of-vocabulary (OOV) character, then $\pvocab(c)$ is zero; similarly, if $c$ does not appear in the source lemma, then $\sum\nolimits_{i: x_i = c} a^t_i$ is zero. The ability to produce OOV characters is one of the primary advantages of pointer-generator models; by contrast models such as our vanilla transformer are restricted to their pre-set vocabulary.

\subsection{Multitask Training}

\begin{figure}[t]
\small
\centering
\begin{tabular}{lllll}
\multirow{ 2}{*}{raw} & grip & grips & V;SG;3;PRS \\
 & grip & gripped & V;PST \\
\midrule
\multirow{ 3}{*}{generated} & grips & grip & V;LEMMA \\
 & grips & gripped & V;PST \\
 & gripped & grip & V;LEMMA \\
\end{tabular}
\caption{English multitask training example (Task 0).}
\label{fig:multitask-exmp}
\end{figure}
Some languages in Task 0 have small training sets, which makes them hard to model. In order to handle that, we perform multitask training, and, thereby, increase the amount of examples available for training.

\paragraph{Morphological reinflection.}
Morphological reinflection is a generalized version of the morphological inflection task, which consists of producing an inflected form for any given source form -- i.e., not necessarily the lemma --, and target tag. For example:

\begin{align}
\textrm{(hugging; V;PST)} \rightarrow \textrm{hugged}.    
\end{align}

 This is a more complex task, since a model needs to infer the underlying lemma of the source form in order to inflect it correctly to the desired form.

Many morphological inflection datasets contain lemmas that are converted to several inflected forms. Treating separate instances for the same source lemma as independent is missing an opportunity to utilize the connection between the different inflected forms.
We approach this by converting our morphological inflection training set into one for morphological reinflection as described in the following. 

\paragraph{From inflection to reinflection.}
Inflected forms of the same lemma are grouped together to sets of one or more (inflected form, morphological features) pairs. Then, for each set, we create new training instances by inflecting all forms to one another, as shown in Figure \ref{fig:multitask-exmp}.
We also let the model inflect forms back to the lemma by adding the lemma as one of the inflected forms, marked with the synthetically generated LEMMA tag. The new training set fully utilizes the connections between different forms in the paradigm, and, in that way, provides more training instances to our model.

\subsection{Hallucination Pretraining}
Another effective tool to improve training in the low-resource setting is data hallucination \citep{DBLP:conf/emnlp/AnastasopoulosN19}. Using hallucination, new pseudo-instances are generated for training, based on suffixation and prefixation rules collected from the original dataset. For languages with less than 1000 training instances, we pretrain our models on a hallucinated training set consisting of 10,000 instances, before training on the multitask training set.

\subsection{Submissions and Ensembling Strategies}
\begin{table}
\centering
\begin{tabular}{lr}
\hline \textbf{Hyperparameter} & \textbf{Value} \\ \hline
Embedding dimension & 256 \\
Encoder layers & 4 \\
Decoder layers & 4 \\
Encoder hidden dimension & 1024 \\
Decoder hidden dimension & 1024 \\
Attention heads & 4 \\
\hline
\end{tabular}
\caption{The hyperparameters used in our inflection models for both Task 0 and Task 2.}
\label{tab:hyperparam-table}
\end{table}

We submit 4 different systems for Task 0. NYU-CUBoulder-2 consists of one pointer-generator transformer model, and, for NYU-CUBoulder-4, we train one vanilla transformer. Those two are our simplest systems and can be seen as baselines for our other submissions. 

Because of the effects of random initialization in non-convex objective functions, we further use ensembling in combination with both architectures: NYU-CUBoulder-1 is an ensemble of three pointer-generator transformers, and NYU-CUBoulder-3 is an ensemble of five pointer-generator transformers. The final decision is made by majority voting. In case of a tie, the answer is chosen randomly among the most frequent predictions. Models participating in the ensembles are from different epochs during the same training run.

As previously stated, all systems are trained on the augmented multitask training sets, and systems trained on languages with less than 1000 training instances were pretrained on the hallucinated datasets.

\subsection{Task 2: Model description}
Our systems for Task 2 consist of a combination of the official baseline system \citep{jin2020unsupervised} and our inflection systems for Task 0.
The system is given raw text and a source file with lemmas, and generates the complete paradigm of each lemma.
The baseline system finds inflected forms in the text,  decides on the number of inflected forms per lemma, and produces pseudo training files for morphological inflection. Any inflections that the system has not found in the raw text are given as test instances. 
Our inflection model then learns from the files and, subsequently, generates all missing forms. We use the pointer-generator and vanilla transformers as our inflection models.

For Task 2, we use ensembling for all submissions.
NYU-CUBoulder-1 is an ensemble of six pointer-generator transformers, NYU-CUBoulder-2 is an ensemble of six vanilla transformers, and NYU-CUBoulder-3 is an ensemble of all twelve models.
For all models in both tasks, we use the  hyperparameters described in Table \ref{tab:hyperparam-table}.

\section{Experiments}

\subsection{Task 0}

\paragraph{Data.}
The dataset for Task 0 covers 90 languages in total: 45 development languages and 45 surprise languages. For  details on the official dataset please refer to \citet{vylomova2020sigmorphon}.

\paragraph{Baselines.}
This year, several baselines are provided for the task. The first system has also been used as a baseline in previous shared tasks on morphological reinflection \citep{DBLP:conf/conll/CotterellKSWVXF17, DBLP:conf/conll/CotterellKSWVMK18}.
It is a non-neural system which first scans the dataset to extract suffix- or prefix-based lemma-to-form transformations. Then, based on the morphological tag at inference time, it applies the most frequent suitable transformation to an input lemma to yield the output form \citep{DBLP:conf/conll/CotterellKSWVXF17}.
The other two baselines are neural models. One is a transformer \citep{DBLP:conf/nips/VaswaniSPUJGKP17,wu2020applying}, and the second one is a hard-attention model \citep{DBLP:conf/acl/WuC19}, which enforces strict monotonicity and learns a latent alignment while learning to transduce. To account for the low-resource settings for some languages, the organizers also employ two additional methods: constructing a multilingual model trained for all languages belonging to each language family \cite{kann-etal-2017-one}, and data augmentation using hallucination \citep{DBLP:conf/emnlp/AnastasopoulosN19}.
Four model types are trained for each neural architecture: a plain model, a family-multilingual model, a data augmented model, and an augmented family-multilingual model.
Overall, there are nine baseline systems for each language.
We compare our models to an oracle baseline by choosing the best score over all baseline systems for each language.
 
\paragraph{Results.}

\begin{table}[t]
\centering
\setlength{\tabcolsep}{4.5pt}
\begin{tabular}{lr|r|r|r|r}
 & Sub-1  & Sub-2 & Sub-3 & Sub-4 & Base\\\hline
\multicolumn{6}{c}{Development Set}\\
\hline
Low & \textbf{88.71} & 88.02 & 84.90 & 84.07 & - \\
Other & 90.46 & 90.63 & 90.20 & \textbf{90.94} & - \\
All  & \textbf{90.06} & 90.02 & 88.96 & 89.34 & - \\
\hline
\multicolumn{6}{c}{Test Set}\\
\hline
Low & 84.8 & 84.8 & 85.5 & 83.9 & \textbf{89.77} \\
Other & 89.7 & 89.8 & 89.8 & 90.2 & \textbf{92.43} \\
All  & 88.6 & 88.7 & 88.8 & 88.8 & \textbf{91.81} \\
\hline
\end{tabular}
\caption{Macro-averaged results over all languages on the official development and test sets for Task 0. Low=languages with less than 1000 train instances, Other=all other languages, All=all languages.}
\label{tab:task0-results}
\end{table}

\begin{table*}[!tb]
\centering
\begin{tabular}{lll|ll|ll|ll|ll}
 System & \multicolumn{ 2}{c}{Baseline 1} & \multicolumn{ 2}{c}{Baseline 2} & \multicolumn{ 2}{c}{Sub-1} & \multicolumn{ 2}{c}{Sub-2} & \multicolumn{ 2}{c}{Sub-3} \\
\hline
\multicolumn{11}{c}{Test Set}\\
\hline
 & slots & macro & slots & macro & slots & macro & slots & macro & slots & macro \\
Basque & 30 & 0.0006 & 27 & 0.0006 & 30 & 0.0005 & 30 & 0.0005 & 30 & \textbf{0.0007} \\
Bulgarian & 35 & 0.283 & 34 & \textbf{0.3169} & 35 & 0.2769 & 35 & 0.2894 & 35 & 0.2789 \\
English & 4 & 0.656 & 4 & \textbf{0.662} & 4 & 0.502 & 4 & 0.528 & 4 & 0.512 \\
Finnish & 21 & 0.0533 & 21 & \textbf{0.055} & 21 & 0.0536 & 21 & 0.0547 & 21 & 0.0535 \\
German & 9 & 0.2835 & 9 & \textbf{0.29} & 9 & 0.273 & 9 & 0.2735 & 9 & 0.2735 \\
Kannada & 172 & 0.1549 & 172 & \textbf{0.1512} & 172 & 0.111 & 172 & 0.1116 & 172 & 0.111 \\
Navajo & 3 & 0.0323 & 3 & \textbf{0.0327} & 3 & 0.004 & 3 & 0.0043 & 3 & 0.0043 \\
Spanish & 29 & 0.2296 & 29 & \textbf{0.2367} & 29 & 0.2039 & 29 & 0.2056 & 29 & 0.203 \\
Turkish & 104 & 0.1421 & 104 & \textbf{0.1553} & 104 & 0.1488 & 104 & 0.1539 & 104 & 0.1513 \\
\textbf{All} &  & 0.2039 &  & \textbf{0.2112} &  & 0.1749 &  & 0.1802 &  & 0.1765 \\
\hline
\end{tabular}
\caption{Results for all test languages on the official test sets for Task 2.}
\label{tab:task2-results-average}
\end{table*}

Our results for Task 0 are displayed in Table \ref{tab:task0-results}. All four systems produce relatively similar results. NYU-CUBoulder-3, our five-model ensemble, performs best overall with $88.8\%$ accuracy on average. We further look at the results for low-resource ($<$ 1000 training examples) and high-resource ($>=$ 1000 training examples) languages separately. This way, we are able to see the advantage of the pointer-generator transformer in the low-resource setting, where all pointer-generator systems achieve an at least $0.9\%$ higher accuracy than the vanilla transformer model. However, in the setting where training data is abundant, the effect of the copy mechanism vanishes, as NYU-CUBoulder-4 -- our only vanilla transformer -- achieved the best results for our high-resource languages.

\subsection{Task 2}

\paragraph{Data.}
For Task 2, a tokenized Bible in each language is given to the participants, along with a list of lemmas. Participants are required to construct the paradigms for all given lemmas.

The languages for Task 2 are again divided into development  and test languages. Development languages are available for model development and hyperparameter tuning, but are not used during the final evaluation. The test languages are used for evaluation only, and do not have development sets. The development languages are: Maltese, Persian, Portuguese, Russian, Swedish. The test languages are: Basque, Bulgarian, English, Finnish, German, Kannada, Navajo, Spanish and Turkish.

\paragraph{Baselines.}
The baseline system for the task is composed of four  components, eventually producing morphological paradigms \cite{jin2020unsupervised}.
The first three modules perform edit tree \citep{Chrupala2020} retrieval, additional lemma retrieval from the corpus, and paradigm size discovery, using distributional information. After the first three steps, pseudo training and test files for morphological inflection are produced.
Finally, the non-neural Task 0 baseline system \cite{DBLP:conf/conll/CotterellKSWVXF17} or the neural transducer by \citet{DBLP:conf/conll/MakarovC18} are used to create missing inflected forms.

\paragraph{Results.}
Systems for Task 2 are evaluated using macro-averaged best-match accuracy \citep{jin2020unsupervised}. Results are shown in in Table \ref{tab:task2-results-average}. 
All three systems produce relatively similar results. NYU-CUBoulder-2, our vanilla transformer ensemble, performed slightly better overall with an average best-match accuracy of $18.02\%$. 
Since our system is close to the baseline models, it performs similarly, achieving slightly worse results. For Basque, our all-round ensemble NYU-CUBoulder-2 outperformed both baselines with a best-match accuracy of $00.07\%$, achieving the highest result in the shared task.

\subsection{Low-resource Setting}
As most inflected forms derive their characters from the source lemma, the use of a mechanism for copying characters directly from the lemma has proven to be effective for morphological inflection generation, especially in the low-resource setting \citep{DBLP:conf/acl/AharoniG17,DBLP:journals/corr/MakarovRC17}.
As all Task 0 datasets are fairly large, we further design a low-resource experiment to investigate the effectiveness of our model.

\begin{table}[b]
\centering
\begin{tabular}{lrrr}
 System & Trm  & Trm-PG & Baseline\\
\hline
All  & 63.06 & 67.61 & \textbf{70.06} \\
\hline
\end{tabular}
\caption{Results on the official development data for our low-resource experiment. Trm=Vanilla transformer, Trm-PG=Pointer-generator transformer, Baseline=neural transducer by \citet{DBLP:conf/conll/MakarovC18}.}
\label{tab:low-res-results-average}
\end{table}
\paragraph{Data.}
We simulate a low-resource setting by sampling 100 instances from all languages that we already consider low-resource, i.e., all languages with less than 1000 training instances. We then keep their development and test sets unchanged. 
Overall, we perform this experiment on 21 languages.

\paragraph{Experimental setup.}
We train a pointer-generator transformer and a vanilla transformer on the modified datasets to examine the effects of the copy mechanism. We keep the hyperparameters unchanged, i.e., they are as mentioned in Table \ref{tab:hyperparam-table}. We use a majority-vote ensemble consisting of 5 individual models for each architecture.

\paragraph{Baseline.}
We additionally train the neural transducer by \citet{DBLP:conf/conll/MakarovC18}, which has achieved the best results for the 2018 shared task in the low-resource setting \citep{DBLP:conf/conll/CotterellKSWVMK18}. The neural transducer uses hard monotonic attention \citep{DBLP:conf/acl/AharoniG17} and transduces the lemma into the inflected form by a sequence of explicit edit operations. It is trained with an imitation learning method \citep{DBLP:conf/conll/MakarovC18}.
We use this model as a reference for the state of the art in the low-resource setting.

\paragraph{Results.}
As seen in Table \ref{tab:low-res-results-average}, for the low-resource dataset, the pointer-generator transformer clearly outperforms the vanilla transformer by an average accuracy of $4.46\%$. For some languages, such as Chichicapan Zapotec, the difference is up to $14\%$. While the neural transducer achieves a higher accuracy, our model performs only $2.45\%$ worse than this state-of-the-art model.\footnote{We could probably obtain better results with appropriate hyperparameter tuning.} We are also able to observe the use of the copy mechanism for copying of OOV characters in the test sets of some languages.

\section{Ablation Studies}

\begin{table}[t]
\centering
\begin{tabular}{lc|c|c|c|c}
Model:
 & 1  & 2 & 3 & 4 & 5 \\
\hline
Copy & \checkmark & \checkmark &  &  & \checkmark \\
Multitask Train & \checkmark &  & \checkmark &  & \checkmark \\
Hallucination  & \checkmark & \checkmark & \checkmark & \checkmark &  \\
\hline
\end{tabular}
\caption{System components for the ablation study for Task 0. Each model is a transformer which contains a combination of the following components: copy mechanism, multitask training and hallucination pretraining.}
\label{tab:ablation-models}
\end{table}

Our systems use three components on top of the vanilla transformer: a copy mechanism, multitask training and hallucination pretraining. We further perform an ablation study to measure the contribution of each component to the overall system performance. For this, we additionally train five different systems with different combinations of components. 
A description of which component is used in which system for this ablation study is shown in Table \ref{tab:ablation-models}.

\subsection{Results}

\paragraph{Copy mechanism.}
Comparing models 2 and 4, which are both trained on the original dataset, pretrained with hallucination and differ only by the use of the copy mechanism, we are able to see that adding this component slightly improves performance by $0.06-0.16\%$. When comparing models 1 and 3, the copy mechanism decreases performance slightly by $0.3\%$ for the high-resource languages and $0.11\%$ overall, but increases performance for low-resource languages by $0.68\%$.

\paragraph{Multitask training.}
Unlike the copy mechanism, multitask training actually consistently decreases the performance of the models. Looking at models 1 and 2, training the pointer-generator transformer on the multitask dataset decreases accuracy by $1.8-2.03\%$ for all three language groups. The same happens for the vanilla transformer with an accuracy decrease of $1.67-2.32\%$. A possible explanation are the relatively large training sets provided for the shared task, as this method is more suitable for the low-resource setting.

\paragraph{Hallucination pretraining.}
In order to examine the effect of hallucination pretraining on our submitted models, we now compare the pointer-generator transformers trained on the multitask data with and without hallucination pretraining (models 1 and 5). Hallucination pretraining shows to be helpful: it increases the accuracy on low-resource languages by $1.85\%$. The performance on the high-resource languages is necessarily the same, as only models for low-resource languages are actually pretrained.

\begin{table}
\centering
\setlength{\tabcolsep}{4.5pt}
\begin{tabular}{lc|c|c|c|c}
{Model:}
 & 1 & 2 & 3 & 4 & 5 \\
\hline
\multicolumn{6}{c}{Development Set}\\
\hline
Low & 88.20 & \textbf{90.00} & 87.52 & 89.84 & 86.35 \\
Other & 90.63 & \textbf{92.66} & 90.93 & 92.60 & 90.63 \\
All  & 90.02 & \textbf{92.04} & 90.13 & 91.96 & 89.63 \\
\hline
\end{tabular}
\caption{Ablation study for Task 0; development set results, averaged over all languages. Low=languages with less than 1000 train instances, Other=all other languages, All=all languages.}
\label{tab:ablation-results}
\end{table}

\section{Conclusion}
We presented the NYU-CUBoulder submissions for SIGMORPHON 2020 Task 0 and Task 2. 

We developed morphological inflection models, based on a transformer and a new model for the task, a pointer-generator transformer, which is a transformer-analogue of a pointer-generator model.
For Task 0, we further added multitask training and hallucination pretraining. 
For Task 2, we combined our inflection models with additional components from the provided baseline  to obtain a fully functional system for unsupervised morphological paradigm completion.

We performed an ablation study to examine the effects of all components of our inflection system. Finally, we designed a low-resource experiment to show that using the copy mechanism on top of the vanilla transformer is beneficial if training sets are small, and achieved results close to a state-of-the-art model for low-resource morphological inflection.

\section*{Acknowledgments}
We would like to thank the organizers of SIGMORPHON 2020 Task 0 and Task 2.

\bibliography{acl2020}

\begin{thebibliography}{32}
\expandafter\ifx\csname natexlab\endcsname\relax\def\natexlab#1{#1}\fi

\bibitem[{Aharoni and Goldberg(2017)}]{DBLP:conf/acl/AharoniG17}
Roee Aharoni and Yoav Goldberg. 2017.
\newblock Morphological inflection generation with hard monotonic attention.
\newblock In \emph{Proceedings of the 55th Annual Meeting of the Association
  for Computational Linguistics, {ACL} 2017, Vancouver, Canada, July 30 -
  August 4, Volume 1: Long Papers}, pages 2004--2015.

\bibitem[{Al{-}Rfou et~al.(2019)Al{-}Rfou, Choe, Constant, Guo, and
  Jones}]{DBLP:conf/aaai/Al-RfouCCGJ19}
Rami Al{-}Rfou, Dokook Choe, Noah Constant, Mandy Guo, and Llion Jones. 2019.
\newblock Character-level language modeling with deeper self-attention.
\newblock In \emph{The Thirty-Third {AAAI} Conference on Artificial
  Intelligence, {AAAI} 2019, The Thirty-First Innovative Applications of
  Artificial Intelligence Conference, {IAAI} 2019, The Ninth {AAAI} Symposium
  on Educational Advances in Artificial Intelligence, {EAAI} 2019, Honolulu,
  Hawaii, USA, January 27 - February 1, 2019.}, pages 3159--3166.

\bibitem[{Anastasopoulos and Neubig(2019)}]{DBLP:conf/emnlp/AnastasopoulosN19}
Antonios Anastasopoulos and Graham Neubig. 2019.
\newblock Pushing the limits of low-resource morphological inflection.
\newblock In \emph{Proceedings of the 2019 Conference on Empirical Methods in
  Natural Language Processing and the 9th International Joint Conference on
  Natural Language Processing, {EMNLP-IJCNLP} 2019, Hong Kong, China, November
  3-7, 2019}, pages 984--996. Association for Computational Linguistics.

\bibitem[{Bahdanau et~al.(2015)Bahdanau, Cho, and
  Bengio}]{DBLP:journals/corr/BahdanauCB14}
Dzmitry Bahdanau, Kyunghyun Cho, and Yoshua Bengio. 2015.
\newblock Neural machine translation by jointly learning to align and
  translate.
\newblock In \emph{3rd International Conference on Learning Representations,
  {ICLR} 2015, San Diego, CA, USA, May 7-9, 2015, Conference Track
  Proceedings}.

\bibitem[{Chrupala(2020)}]{Chrupala2020}
Grzegorz Chrupala. 2020.
\newblock Towards a machine-learning architecture for lexical functional
  grammar parsing.

\bibitem[{Cotterell et~al.(2018)Cotterell, Kirov, Sylak{-}Glassman, Walther,
  Vylomova, McCarthy, Kann, Mielke, Nicolai, Silfverberg, Yarowsky, Eisner, and
  Hulden}]{DBLP:conf/conll/CotterellKSWVMK18}
Ryan Cotterell, Christo Kirov, John Sylak{-}Glassman, G{\'{e}}raldine Walther,
  Ekaterina Vylomova, Arya~D. McCarthy, Katharina Kann, S.~J. Mielke, Garrett
  Nicolai, Miikka Silfverberg, David Yarowsky, Jason Eisner, and Mans Hulden.
  2018.
\newblock The conll-sigmorphon 2018 shared task: Universal morphological
  reinflection.
\newblock In \emph{Proceedings of the CoNLL {SIGMORPHON} 2018 Shared Task:
  Universal Morphological Reinflection, Brussels, October 31 - November 1,
  2018}, pages 1--27. Association for Computational Linguistics.

\bibitem[{Cotterell et~al.(2017)Cotterell, Kirov, Sylak{-}Glassman, Walther,
  Vylomova, Xia, Faruqui, K{\"{u}}bler, Yarowsky, Eisner, and
  Hulden}]{DBLP:conf/conll/CotterellKSWVXF17}
Ryan Cotterell, Christo Kirov, John Sylak{-}Glassman, G{\'{e}}raldine Walther,
  Ekaterina Vylomova, Patrick Xia, Manaal Faruqui, Sandra K{\"{u}}bler, David
  Yarowsky, Jason Eisner, and Mans Hulden. 2017.
\newblock Conll-sigmorphon 2017 shared task: Universal morphological
  reinflection in 52 languages.
\newblock In \emph{Proceedings of the CoNLL {SIGMORPHON} 2017 Shared Task:
  Universal Morphological Reinflection, Vancouver, BC, Canada, August 3-4,
  2017}, pages 1--30. Association for Computational Linguistics.

\bibitem[{Cotterell et~al.(2016{\natexlab{a}})Cotterell, Kirov,
  Sylak{-}Glassman, Yarowsky, Eisner, and
  Hulden}]{DBLP:conf/sigmorphon/CotterellKSYEH16}
Ryan Cotterell, Christo Kirov, John Sylak{-}Glassman, David Yarowsky, Jason
  Eisner, and Mans Hulden. 2016{\natexlab{a}}.
\newblock The {SIGMORPHON} 2016 shared task - morphological reinflection.
\newblock In \emph{Proceedings of the 14th {SIGMORPHON} Workshop on
  Computational Research in Phonetics, Phonology, and Morphology, Berlin,
  Germany, August 11, 2016}, pages 10--22. Association for Computational
  Linguistics.

\bibitem[{Cotterell et~al.(2016{\natexlab{b}})Cotterell, Sch{\"{u}}tze, and
  Eisner}]{DBLP:conf/acl/CotterellSE16}
Ryan Cotterell, Hinrich Sch{\"{u}}tze, and Jason Eisner. 2016{\natexlab{b}}.
\newblock Morphological smoothing and extrapolation of word embeddings.
\newblock In \emph{Proceedings of the 54th Annual Meeting of the Association
  for Computational Linguistics, {ACL} 2016, August 7-12, 2016, Berlin,
  Germany, Volume 1: Long Papers}. The Association for Computer Linguistics.

\bibitem[{Devlin et~al.(2019)Devlin, Chang, Lee, and
  Toutanova}]{DBLP:conf/naacl/DevlinCLT19}
Jacob Devlin, Ming{-}Wei Chang, Kenton Lee, and Kristina Toutanova. 2019.
\newblock {BERT:} pre-training of deep bidirectional transformers for language
  understanding.
\newblock In \emph{Proceedings of the 2019 Conference of the North American
  Chapter of the Association for Computational Linguistics: Human Language
  Technologies, {NAACL-HLT} 2019, Minneapolis, MN, USA, June 2-7, 2019, Volume
  1 (Long and Short Papers)}, pages 4171--4186. Association for Computational
  Linguistics.

\bibitem[{Erdmann et~al.(2020)Erdmann, Elsner, Wu, Cotterell, and
  Habash}]{DBLP:journals/corr/abs-2005-01630}
Alexander Erdmann, Micha Elsner, Shijie Wu, Ryan Cotterell, and Nizar Habash.
  2020.
\newblock The paradigm discovery problem.
\newblock \emph{CoRR}, abs/2005.01630.

\bibitem[{Gorman et~al.(2020)Gorman, Ashby, Goyzueta, McCarthy, Wu, and
  You}]{Task1-2020-sigmorphon}
Kyle Gorman, Lucas~F.E. Ashby, Aaron Goyzueta, Arya~D. McCarthy, Shijie Wu, and
  Daniel You. 2020.
\newblock The sigmorphon 2020 shared task on multilingual grapheme-to-phoneme
  conversion.
\newblock In \emph{Proceedings of the 17th {SIGMORPHON} Workshop on
  Computational Research in Phonetics, Phonology, and Morphology}. Association
  for Computational Linguistics.

\bibitem[{Hochreiter and Schmidhuber(1997)}]{DBLP:journals/neco/HochreiterS97}
Sepp Hochreiter and J{\"{u}}rgen Schmidhuber. 1997.
\newblock Long short-term memory.
\newblock \emph{Neural Computation}, 9(8):1735--1780.

\bibitem[{Janecki(2000)}]{Janecki1998}
Klara Janecki. 2000.
\newblock 300 polish verbs.
\newblock Barron’s Educational Series.

\bibitem[{Jin et~al.(2020)Jin, Cai, Peng, Xia, McCarthy, and
  Kann}]{jin2020unsupervised}
Huiming Jin, Liwei Cai, Yihui Peng, Chen Xia, Arya~D. McCarthy, and Katharina
  Kann. 2020.
\newblock Unsupervised morphological paradigm completion.
\newblock In \emph{Proceedings of the 58th Annual Meeting of the Association
  for Computational Linguistics}. Association for Computational Linguistics.

\bibitem[{Kann et~al.(2017)Kann, Cotterell, and
  Sch{\"u}tze}]{kann-etal-2017-one}
Katharina Kann, Ryan Cotterell, and Hinrich Sch{\"u}tze. 2017.
\newblock One-shot neural cross-lingual transfer for paradigm completion.
\newblock In \emph{Proceedings of the 55th Annual Meeting of the Association
  for Computational Linguistics (Volume 1: Long Papers)}, pages 1993--2003,
  Vancouver, Canada. Association for Computational Linguistics.

\bibitem[{Kann et~al.(2020)Kann, McCarthy, Nicolai, and
  Hulden}]{kann-etal-2020-sigmorphon}
Katharina Kann, Arya~D. McCarthy, Garrett Nicolai, and Mans Hulden. 2020.
\newblock The {SIGMORPHON} 2020 shared task on unsupervised morphological
  paradigm completion.
\newblock In \emph{Proceedings of the 17th SIGMORPHON Workshop on Computational
  Research in Phonetics, Phonology, and Morphology}. Association for
  Computational Linguistics.

\bibitem[{Kann and
  Sch{\"{u}}tze(2016{\natexlab{a}})}]{DBLP:conf/sigmorphon/KannS16}
Katharina Kann and Hinrich Sch{\"{u}}tze. 2016{\natexlab{a}}.
\newblock {MED:} the {LMU} system for the {SIGMORPHON} 2016 shared task on
  morphological reinflection.
\newblock In \emph{Proceedings of the 14th {SIGMORPHON} Workshop on
  Computational Research in Phonetics, Phonology, and Morphology, Berlin,
  Germany, August 11, 2016}, pages 62--70.

\bibitem[{Kann and Sch{\"{u}}tze(2016{\natexlab{b}})}]{DBLP:conf/acl/KannS16}
Katharina Kann and Hinrich Sch{\"{u}}tze. 2016{\natexlab{b}}.
\newblock Single-model encoder-decoder with explicit morphological
  representation for reinflection.
\newblock In \emph{Proceedings of the 54th Annual Meeting of the Association
  for Computational Linguistics, {ACL} 2016, August 7-12, 2016, Berlin,
  Germany, Volume 2: Short Papers}.

\bibitem[{Kibrik(1998)}]{kibrik1998archi}
Aleksandr~E. Kibrik. 1998.
\newblock The handbook of morphology.
\newblock In \emph{Andrew Spencer and Arnold~M. Zwicky, editors}, pages
  455--476. Oxford: Blackwell Publishers.

\bibitem[{Makarov and Clematide(2018)}]{DBLP:conf/conll/MakarovC18}
Peter Makarov and Simon Clematide. 2018.
\newblock {UZH} at conll-sigmorphon 2018 shared task on universal morphological
  reinflection.
\newblock In \emph{Proceedings of the CoNLL {SIGMORPHON} 2018 Shared Task:
  Universal Morphological Reinflection, Brussels, October 31 - November 1,
  2018}, pages 69--75. Association for Computational Linguistics.

\bibitem[{Makarov et~al.(2017)Makarov, Ruzsics, and
  Clematide}]{DBLP:journals/corr/MakarovRC17}
Peter Makarov, Tatiana Ruzsics, and Simon Clematide. 2017.
\newblock Align and copy: {UZH} at {SIGMORPHON} 2017 shared task for
  morphological reinflection.
\newblock \emph{CoRR}, abs/1707.01355.

\bibitem[{Minkov et~al.(2007)Minkov, Toutanova, and
  Suzuki}]{DBLP:conf/acl/MinkovTS07}
Einat Minkov, Kristina Toutanova, and Hisami Suzuki. 2007.
\newblock Generating complex morphology for machine translation.
\newblock In \emph{{ACL} 2007, Proceedings of the 45th Annual Meeting of the
  Association for Computational Linguistics, June 23-30, 2007, Prague, Czech
  Republic}. The Association for Computational Linguistics.

\bibitem[{See et~al.(2017)See, Liu, and Manning}]{DBLP:conf/acl/SeeLM17}
Abigail See, Peter~J. Liu, and Christopher~D. Manning. 2017.
\newblock Get to the point: Summarization with pointer-generator networks.
\newblock In \emph{Proceedings of the 55th Annual Meeting of the Association
  for Computational Linguistics, {ACL} 2017, Vancouver, Canada, July 30 -
  August 4, Volume 1: Long Papers}, pages 1073--1083.

\bibitem[{Seeker and {\c{C}}etinoglu(2015)}]{DBLP:journals/tacl/SeekerC15}
Wolfgang Seeker and {\"{O}}zlem {\c{C}}etinoglu. 2015.
\newblock A graph-based lattice dependency parser for joint morphological
  segmentation and syntactic analysis.
\newblock \emph{Trans. Assoc. Comput. Linguistics}, 3:359--373.

\bibitem[{Sharma et~al.(2018)Sharma, Katrapati, and
  Sharma}]{DBLP:conf/conll/SharmaKS18}
Abhishek Sharma, Ganesh Katrapati, and Dipti~Misra Sharma. 2018.
\newblock {IIT(BHU)-IIITH} at conll-sigmorphon 2018 shared task on universal
  morphological reinflection.
\newblock In \emph{Proceedings of the CoNLL {SIGMORPHON} 2018 Shared Task:
  Universal Morphological Reinflection, Brussels, October 31 - November 1,
  2018}, pages 105--111.

\bibitem[{Vaswani et~al.(2017)Vaswani, Shazeer, Parmar, Uszkoreit, Jones,
  Gomez, Kaiser, and Polosukhin}]{DBLP:conf/nips/VaswaniSPUJGKP17}
Ashish Vaswani, Noam Shazeer, Niki Parmar, Jakob Uszkoreit, Llion Jones,
  Aidan~N. Gomez, Lukasz Kaiser, and Illia Polosukhin. 2017.
\newblock Attention is all you need.
\newblock In \emph{Advances in Neural Information Processing Systems 30: Annual
  Conference on Neural Information Processing Systems 2017, 4-9 December 2017,
  Long Beach, CA, {USA}}, pages 5998--6008.

\bibitem[{Vinyals et~al.(2015)Vinyals, Kaiser, Koo, Petrov, Sutskever, and
  Hinton}]{DBLP:conf/nips/VinyalsKKPSH15}
Oriol Vinyals, Lukasz Kaiser, Terry Koo, Slav Petrov, Ilya Sutskever, and
  Geoffrey~E. Hinton. 2015.
\newblock Grammar as a foreign language.
\newblock In \emph{Advances in Neural Information Processing Systems 28: Annual
  Conference on Neural Information Processing Systems 2015, December 7-12,
  2015, Montreal, Quebec, Canada}, pages 2773--2781.

\bibitem[{Vylomova et~al.(2020)Vylomova, White, Salesky, Mielke, Wu, Ponti,
  Maudslay, Zmigrod, Valvoda, Toldova, Tyers, Klyachko, Yegorov, Krizhanovsky,
  Czarnowska, Nikkarinen, Krizhanovsky, Pimentel, Hennigen, Kirov, Nicolai,
  Williams, Anastasopoulos, Cruz, Chodroff, Cotterell, Silfverberg, and
  Hulden}]{vylomova2020sigmorphon}
Ekaterina Vylomova, Jennifer White, Elizabeth Salesky, Sabrina~J. Mielke,
  Shijie Wu, Edoardo Ponti, Rowan~Hall Maudslay, Ran Zmigrod, Joseph Valvoda,
  Svetlana Toldova, Francis Tyers, Elena Klyachko, Ilya Yegorov, Natalia
  Krizhanovsky, Paula Czarnowska, Irene Nikkarinen, Andrej Krizhanovsky, Tiago
  Pimentel, Lucas~Torroba Hennigen, Christo Kirov, Garrett Nicolai, Adina
  Williams, Antonios Anastasopoulos, Hilaria Cruz, Eleanor Chodroff, Ryan
  Cotterell, Miikka Silfverberg, and Mans Hulden. 2020.
\newblock The {SIGMORPHON 2020 Shared Task 0}: Typologically diverse
  morphological inflection.
\newblock In \emph{Proceedings of the 17th SIGMORPHON Workshop on Computational
  Research in Phonetics, Phonology, and Morphology}.

\bibitem[{Wu and Cotterell(2019)}]{DBLP:conf/acl/WuC19}
Shijie Wu and Ryan Cotterell. 2019.
\newblock Exact hard monotonic attention for character-level transduction.
\newblock In \emph{Proceedings of the 57th Conference of the Association for
  Computational Linguistics, {ACL} 2019, Florence, Italy, July 28- August 2,
  2019, Volume 1: Long Papers}, pages 1530--1537. Association for Computational
  Linguistics.

\bibitem[{Wu et~al.(2020)Wu, Cotterell, and Hulden}]{wu2020applying}
Shijie Wu, Ryan Cotterell, and Mans Hulden. 2020.
\newblock Applying the transformer to character-level transduction.

\bibitem[{Zhao et~al.(2019)Zhao, Wang, Shen, Jia, and
  Liu}]{DBLP:conf/naacl/ZhaoWSJL19}
Wei Zhao, Liang Wang, Kewei Shen, Ruoyu Jia, and Jingming Liu. 2019.
\newblock Improving grammatical error correction via pre-training a
  copy-augmented architecture with unlabeled data.
\newblock In \emph{Proceedings of the 2019 Conference of the North American
  Chapter of the Association for Computational Linguistics: Human Language
  Technologies, {NAACL-HLT} 2019, Minneapolis, MN, USA, June 2-7, 2019, Volume
  1 (Long and Short Papers)}, pages 156--165.

\end{thebibliography}
\bibliographystyle{acl_natbib}

\end{document}